\documentclass[letter, 10 pt, conference]{ieeeconf}  

\IEEEoverridecommandlockouts                              

\usepackage{amsmath}
\usepackage{amssymb}
\usepackage{mathtools}
\usepackage{graphicx}
\usepackage{float}
\usepackage{array}
\usepackage{listings}
\usepackage{graphicx}
\usepackage{cite}
\usepackage{dsfont}
\usepackage{mathtools,etoolbox}
\usepackage[ruled, linesnumbered]{algorithm2e}
\usepackage[none]{hyphenat}
\usepackage[utf8]{inputenc}
\usepackage[english]{babel}
\usepackage[T1]{fontenc}

\usepackage{amsthm}
\usepackage{xfrac}
\usepackage{nicefrac}
\usepackage{booktabs}
\usepackage{flushend}
\usepackage[switch, pagewise]{lineno}
\usepackage{mathrsfs}
\DeclareMathAlphabet{\mathpzc}{OT1}{pzc}{m}{it}
\usepackage{multirow}
\usepackage{multicol}
\usepackage{bbm}
\usepackage{soul}
\usepackage{xfrac}
\usepackage{balance}
\usepackage{titlesec}
\usepackage{caption}
\usepackage{subcaption}
\usepackage[final]{pdfpages}
\usepackage{hyperref}
\usepackage{resizegather}
\usepackage{svg}
\usepackage{ctable}
\usepackage{cite}
\usepackage{xcolor}%

\hypersetup{colorlinks=true,urlcolor=blue,citecolor=red}

\title{\LARGE \bf \textit{VizFlyt:} Perception-centric Pedagogical Framework For Autonomous Aerial Robots}

\author{\normalsize{Kushagra Srivastava$^{1,*}$, Rutwik Kulkarni$^{1,*}$, Manoj Velmurugan$^{1,*}$, Nitin J. Sanket$^{1}$}%
\thanks{\textit{$^*$ Equal Contribution, author order decided at random. Corresponding author: Manoj Velmurugan} (\texttt{mvelmurugan@wpi.edu}). $^1$Perception and Autonomous Robotics (PeAR) Group, Worcester Polytechnic Institute. This work is partly supported by the National Science Foundation I/UCRC grant with number 1939061 with support from RTX Technology Research Center.}}

\begin{document}
\makeatletter
\g@addto@macro\@maketitle{
\begin{figure}[H]
   \setlength{\linewidth}{\textwidth}
   \setlength{\hsize}{\textwidth}
    \centering
    \includegraphics[width=\textwidth]{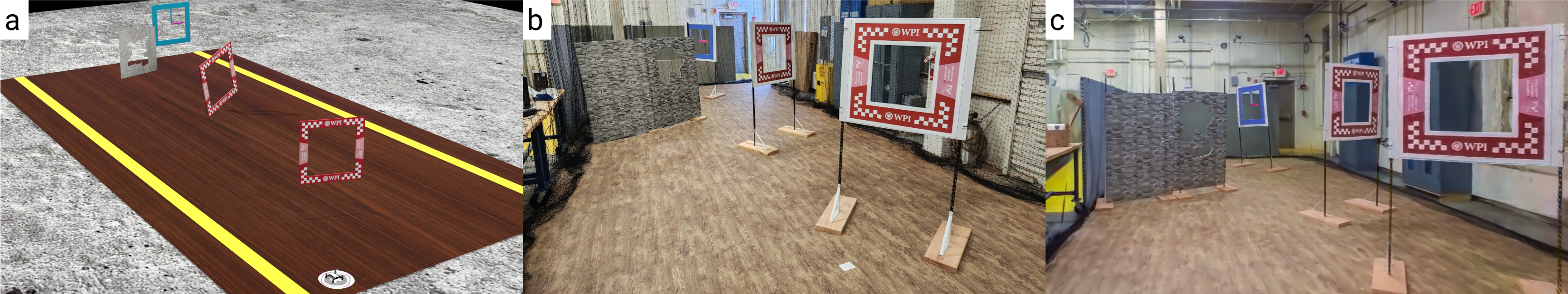}
    \caption{Evolution of perception-centric offerings of the course. (a) course based on simulation, (b) course based on a real-world obstacle course, and (c) new course based on the proposed \textit{VizFlyt} framework, where the image is a real-time photorealistic render of a hallucinated camera on an aerial robot used for autonomy tasks. \textit{All the images in this paper are best viewed in color on a computer screen at 200\% zoom.}}
    \vspace{-20pt}
    \label{fig:Overview}
    \end{figure}
}
\maketitle
\setcounter{figure}{1} 

\begin{abstract}
Autonomous aerial robots are becoming commonplace in our lives. Hands-on aerial robotics courses are pivotal in training the next-generation workforce to meet the growing market demands. Such an efficient and compelling course depends on a reliable testbed. In this paper, we present \textit{VizFlyt}, an open-source perception-centric Hardware-In-The-Loop  (HITL) photorealistic testing framework for aerial robotics courses. We utilize pose from an external localization system to hallucinate real-time and photorealistic visual sensors using 3D Gaussian Splatting. This enables stress-free testing of autonomy algorithms on aerial robots without the risk of crashing into obstacles. We achieve over 100Hz of system update rate. Lastly, we build upon our past experiences of offering hands-on aerial robotics courses and propose a new open-source and open-hardware curriculum based on \textit{VizFlyt} for the future.  We test our framework on various course projects in real-world HITL experiments and present the results showing the efficacy of such a system and its large potential use cases. Code, datasets, hardware guides and demo videos are available at \href{https://pear.wpi.edu/research/vizflyt.html}{https://pear.wpi.edu/research/vizflyt.html}

\end{abstract}


\section{Introduction}

The use of aerial robots (drones) has surged in the past decade, with applications ranging from search and rescue \cite{search_rescue} to farming \cite{farming} and inspection \cite{inspection}. However, most drones are manually operated, limiting their scalability and use in harsh conditions.  Hence, there is a growing need for autonomous drones and the software to operate them. Aerial robotics courses \cite{pirdone, mit_course, penn_course, tum_course, umd_course, rbe595, meam_penn_course} play a pivotal role in training students from K-12 to graduate levels, focusing on skills from piloting to autonomy software development. While many courses teach manual piloting \cite{osu, osu1} or drone hardware building \cite{pirdone}, fewer focus on real-time autonomy using onboard sensing and computation. Hands-on learning has consistently proven effective for robotics education, and aerial robotics is no exception. However, running a hands-on drone autonomy course faces challenges such as (a) procuring the right hardware platform with long-term support, (b) ensuring safe operation for students, (c) minimizing costs for operation, maintenance, and repair, (d) providing an extensive and user-friendly API for sensor and control access, (e) allowing modularity for adding new sensor/computational payloads, and (f) ensuring stability, robustness, and repeatability. Depending on the course's focus, some factors may outweigh others. For example, in courses focused on teaching controls and planning, it might be possible to avoid using onboard sensors and simply use an external localization system mated to an offboard PC that communicates to the robot via a wireless link. Furthermore, in courses that involve sensor fusion or perception, it becomes essential to use onboard sensors/computation due to the requirement of high rate and low latency processing to hit the control targets required.
This paper focuses on developing a scalable, robust framework for teaching vision-based drone autonomy, which is widely used for its cost-effectiveness and ubiquity.

Currently, aerial robotics courses rely on custom-built or educational drone platforms \cite{mambo, djitello, rollingspider}, many of which are discontinued. This presents significant logistical challenges, as course materials must be regularly updated to accommodate new hardware. The issue is compounded by hardware damage from collisions with environmental elements, leading to costly spare part replacements. Additionally, an autonomy-focused aerial robotics course requires integrating hardware like front-facing, downward-facing, and/or stereo cameras. This added complexity increases both the effort and cost of running the course, making it difficult to scale for larger student groups.

A solution is to conduct the course entirely on a simulator\cite{flightmare, Gazebo, IsaacSim}. We encourage reading \cite{dimmig2024survey} for a study on aerial robot simulators. While simulators can effectively replicate certain aspects of drone operation, they fall short in providing the real-world experience needed to develop robust, reliable algorithms for safety-critical systems in aerial vehicles. 


Recent research \cite{agilicious, FlightGoggles_MIT, song2023learning} has utilized Hardware-In-The-Loop (HITL) simulation, where a physical robot flies in the real world, eliminating the need to simulate flight dynamics. This approach is more flexible than using real robots with onboard sensing and computation, as it allows for quick adaptation to various environmental settings, reduces collision risk, and retains flight dynamics fidelity. Inspired by this, we combine a photorealistic rendering framework with HITL to create a scalable method for teaching. This enables students to focus on autonomy concepts at any level (perception, planning, and controls) in a realistic environment. For instructors and teaching assistants, it reduces the logistical burden (e.g., maintaining drones and experimental setups), lowers operational costs, minimizes repair times, and simplifies live-grading efforts.


The authors have collectively participated in five aerial robotics courses \cite{meam_penn_course, umd_course, rbe595} across three universities (University of Pennsylvania, University of Maryland, College Park, and Worcester Polytechnic Institute) in roles ranging from student to teaching assistant to instructor. These courses spanned multiple departments (Robotics Engineering, Computer Science, Aerospace Engineering, and Mechanical Engineering) and catered to both undergraduate and graduate students. Our experiences have provided us with a well-rounded perspective on running a hands-on drone autonomy course with minimal logistical challenges.

We propose \textit{VizFlyt}, a perception-centric pedagogical framework for autonomous aerial robotics based on HITL concepts. By leveraging 3D Gaussian Splatting (3DGS) \cite{kerbl3Dgaussians} for real-time view synthesis from robot pose, \textit{VizFlyt} decouples aerial robotics coursework from hardware dependency, providing a flexible, robust, and cost-effective solution for testing in complex visual environments. It is an intuitive framework designed for ease of use, expandable to novel features, requiring no prior background knowledge, making it a perfect framework for teaching aerial robotics. The key contributions of the \textit{VizFlyt} framework are:

\begin{itemize} \item \textit{VizFlyt} eliminates the need for onboard proprioceptive sensors, reducing hardware complexity and reliance on specific robots, while facilitating RGBD updates at over 100 Hz per sensor. \item \textit{VizFlyt} enables rapid dynamic asset generation for novel environment creation using synthetic or real data, supporting flight testing across a variety of scenarios. \item We demonstrate \textit{VizFlyt}'s effectiveness for teaching aerial robotics through key tasks such as Visual Odometry (VO), high-speed obstacle avoidance, and navigation through known and unknown gaps in real-world experiments. \item The \textit{VizFlyt} framework (software and hardware) will be open-source, making it future-proof and expandable. \end{itemize}

\section{Learnings From Current Curricula}
\subsection{Others' Aerial Robotics Coursework}

Several aerial robotics courses offer first-hand experience and often involve 3 distinctive modules (i) Development and assembly of drone hardware, (ii) Flying and operating drones, (iii) Development of drone autonomy software. Drone autonomy encompasses tasks or objectives focused on motion planning \cite{meam_penn_course, penn_course, umd_course}, control systems \cite{meam_penn_course, pirdone, tum_course, umd_course}, and perception algorithms \cite{meam_penn_course, mit_course, pirdone, tum_course, umd_course}. Course projects are tailored around fundamental tasks such as SLAM, sensor fusion, quadrotor dynamics, path planning, pose estimation, and obstacle avoidance. While some of the projects provide an introductory understanding of quadrotor operation, others are more advanced, catering to graduate-level expertise. Some of the courses focus on software integration for robust performance. However, testing under various conditions of these projects is limited due to time constraints, and dynamic variations in the environment are not addressed, as static datasets are used to facilitate easier evaluation. Majority of these courses use educational drones \cite{djitello, mambo, rollingspider} that are currently discontinued rendering the course inoperable.

\subsection{Our Latest Past Course Offering}
The aerial robotics courses we have taught \cite{yiannis, umd_course, fire, rbe595} have been built around the philosophy of gamification of education (Fig. \ref{fig:Overview}), where advanced course knowledge is disseminated to students by encouraging healthy competition to learn quickly, particularly useful for aerial robotics with a high barrier of entry.

\href{https://pear.wpi.edu/teaching/rbe595/fall2023.html}{RBE595-F02-ST: Hands-On Autonomous Aerial Robotics} is an advanced graduate course tailored for robotics students. The course dealt with advanced concepts of vision-based autonomy for a challenging obstacle course based on multiple drone racing competitions \cite{foehn2022alphapilot, motorsport_concept, drl_website}. Specifically, the course included flight through drone-racing like windows, an unknown-shaped gap and a dynamic window. The course was run using a custom Blender$^\text{\textregistered}$ simulator with dynamics mimicking the ArduCopter stack for the initial projects and the final projects were performed on DJI Tello Edu drones. Over the course, 4/6 teams completed the final obstacle course. Though the course ran successfully, it had challenges which need to be addressed for the course to smoothly run in the future.   

\textbf{Instructor and Teaching Assistant Feedback:} The major setback in the course was frequent failure of drone hardware with about eight robots, multiple propellers, multiple propeller guards, 10 batteries breaking completely due to wear and tear and crashes. This necessitated keeping a stockpile of spare drones (parts), which was burdening and increased the cost of operation. The DJI Tello Edu was discontinued after the first run of this course which forces a redesign of the course. Furthermore, grading using live demonstrations took significant time since autograding was not possible. This coupled with scheduling conflicts for evaluation resulted in a rigorous responsibility.

\textbf{Student Feedback:}
The majority of students pointed to hardware reliability as their biggest issue which included unreliable takeoffs, frequent connection problems and packet loss issues. The Python API for controlling the drone was seen as cumbersome and inconsistent. On average, students reported crashing the drone more than 20 times per project. 
The students reported that they spent a significant amount of time dealing with hardware issues (about 40\% of the total time spent), rather than solving the actual project.

Overall, looking back from the first offering in 2017\cite{yiannis} to the latest offering in 2023\cite{rbe595}, we learned that educational drones constantly get discontinued, resulting in a lot of breakages due to collisions with the environmental components of the obstacle course, present major safety concerns making recurring offerings of the courses hard due to limited budgets, smaller offering capacity and lack of enough assistant support to maintain robots. To this end, we noticed that we have to solve the following problems: (a) minimize environmental components in the flying space to decrease the probability of collision, (b) change to open-source hardware and software designs to enable future support and adaptation, (c) robust robot design to minimize breakages, (d) scalability to various environments that can be photo-realistically rendered in real-time for vision-based autonomy. The \textit{VizFlyt} framework presented next will be used as the base of the proposed course curriculum to address the aforementioned issues.

\section{\textit{VizFlyt} Framework}
\subsection{Digital Twin Scene Generation}
The first step is to create a digital model or clone of the real world to be used as the base of HITL. This can be obtained either from digital modeling software like Blender$^\text{\textregistered}$ where assets (for obstacles) are manually crafted or obtained through photogrammetry, which is generally time-consuming. As an efficient alternative, we train high-fidelity 3DGS models, that represent complete scenes (not just the obstacles), using real images which will be later used for hallucinating proprioceptive vision-based sensors. We ensure that rendering from the 3DGS scene is consistent across viewpoints by capturing images from diverse perspectives.



\subsection{Hardware Setup}


To enable accurate HITL testing for autonomy stack, an aerial robot flies in a netted flying space equipped with an external localization system such as Vicon$^\text{\textregistered}$ motion capture\cite{vicon}, Bitcraze LoCo.\cite{bitcraze_loco}, April Tags\cite{DBLP:conf/icra/PfrommerSDC17} or IR based tracking\cite{lighthouse}, providing accurate real-time robot pose updates. 
The robot has an onboard IMU, but all other visual sensors such as RGB or RGBD cameras are synthesized (hallucinated) using 3DGS from the external pose (akin to Virtual Reality for drones). The real and hallucinated sensor data are used for autonomy processing on a tabletop embedded computer (to simulate real-time constraints). 


\begin{figure}[b!]
    \centering
    \includegraphics[width=\columnwidth]{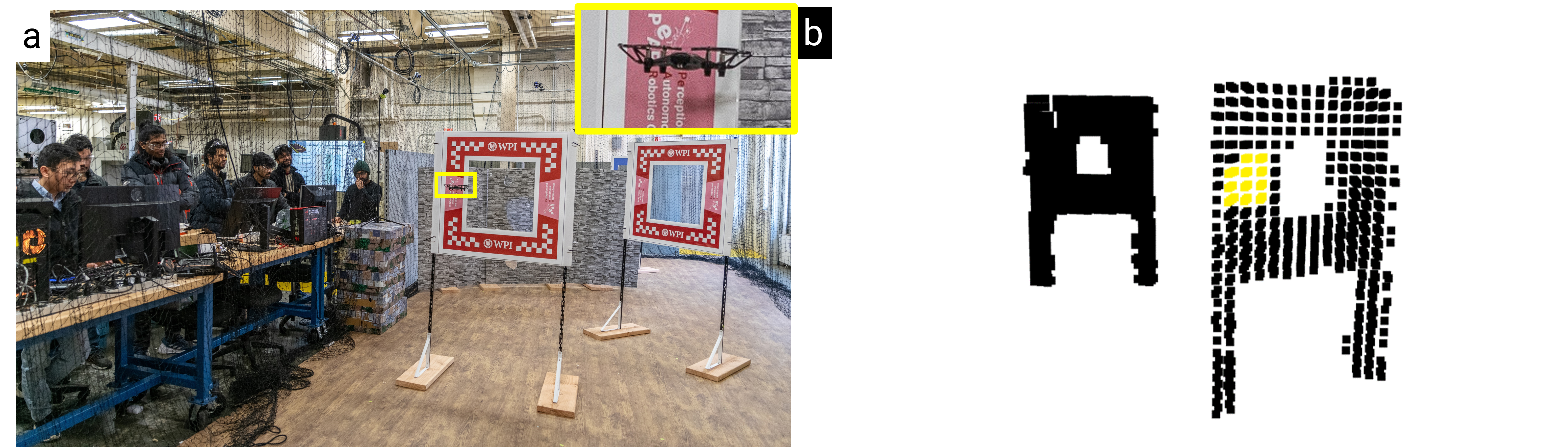}
    \caption{Termination of attempt during a crash (yellow highlight). (a) Real crash in the old version of the course leading to damages and manual grading, (b) automatic crash reporting from the hallucinated camera with no robot damages. }
    \label{fig:collision_check}
\end{figure}

\subsection{Collision Detection}\label{sec:collisioncheck}
To detect collisions for auto-grading assignments, a robust collision detection system is required. Given that the 3DGS environment does not include built-in collision detection methods, we calculate a 3D occupancy voxel grid map by exporting the 3DGS environment as a point cloud. A radius-based collision check is employed where we check for voxels lying inside a fixed-radius sphere centered around the current robot position. When a collision is detected in the 3DGS environment, the real-world drone is promptly switched to land mode with a visualizer indication (Fig. \ref{fig:collision_check}).

\subsection{Framework Architecture} \label{sec:framework_architecture}

\begin{figure}[b!]
    \centering
    \includegraphics[width=\columnwidth]{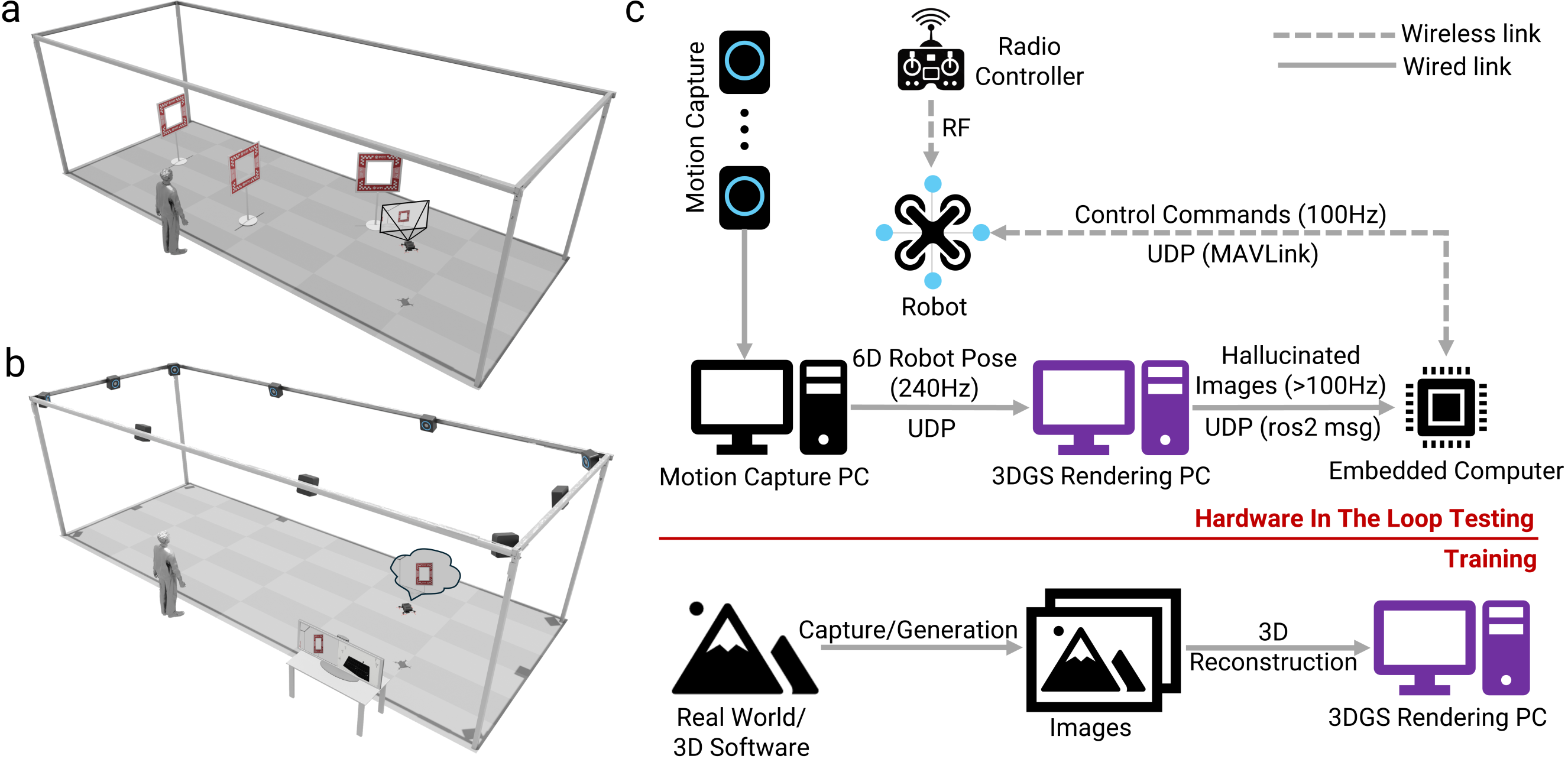}
    \caption{(a) Artistic impression of the previous iteration of the course with experiments run with real obstacles and sensors, (b) Artistic impression of the proposed course with experiments run using the \textit{VizFlyt} framework to hallucinate sensors for HITL testing, and (c) \textit{VizFlyt} framework Architecture. }
    \label{fig:framework_architecture}
\end{figure}

We use 14 Vicon$^\text{\textregistered}$ Vero V2.2 cameras to obtain the drone's real-time pose on a Windows PC (Motion Capture PC, Fig. \ref{fig:framework_architecture}). This pose data is then transferred via UDP to a ROS2 (Humble) \cite{ros2} node running on a dedicated Ubuntu 22.04 machine (3DGS Rendering PC with an Intel$^\text{\textregistered}$ Core i7 processor with  48GB RAM and NVIDIA$^\text{\textregistered}$ RTX 3080 GPU with 10GB VRAM). The 3DGS Rendering PC is responsible for generating a pose-dependent RGBD stream which is published via a ROS2 node. 

To simulate real-time hardware constraints, we implement our vision-based autonomy software on an embedded platform, such as the NVIDIA Jetson Orin Nano. The embedded board is connected to the rendering PC via Ethernet and RGBD frames are obtained by subscribing to the ROS2 node.

After the frames are processed, the control commands are computed and transmitted to the robot using the MAVLink UDP protocol over an ESP8266-based Wi-Fi link. Our framework remains hardware-agnostic up to the drone protocol level, allowing compatibility with a variety of systems. 
To minimize crashes, geo-fencing is implemented using the ground truth pose of the drone, ensuring it lands automatically if it exceeds the boundaries of the designated safe flying area.







\section{Proposed Course Curriculum} \label{sec:proposed_course_curriculum}
The next offering of the course will be based on the \textit{VizFlyt} framework as described next.

\subsection{Learning Objectives}
The course is designed as a senior undergraduate class and/or an advanced graduate class. The course will introduce students to the fundamental principles and challenges of real-time perception and autonomy in aerial robotics. It aims to provide a hands-on experience with high-level tasks such as object detection, visual servoing, depth estimation, obstacle avoidance, and navigation using onboard real-time constraints of sensing and computation. The curriculum focuses on developing the students' ability to design, implement, and evaluate algorithms that enable autonomous aerial robots to interact with their environments in complex conditions. At the culmination of the course, students are expected to understand both theoretical and practical aspects of autonomous navigation, as well as develop skills that are directly applicable to real-world aerial robotics challenges. 

\subsection{Course Setup}
The course consists of a custom Blender$^\text{\textregistered}$ based simulator with quadrotor dynamics and the control stack build around Arducopter to test the initial prototypes of algorithms and generation of data for deep learning training. For the hardware experiments, we use a custom-built quadrotor platform PeARWhippet160 in the \textit{VizFlyt} framework as described in $\S$\ref{sec:framework_architecture}. We use the information from onboard IMU and hallucinated RGB/RGBD sensors. Different sensor suites can be used to adapt the difficulty of the projects in the course for various educational levels and technical backgrounds, making our curriculum widely applicable.

\subsection{Course Curriculum}
The course is semester-long hands-on and project based with six projects and 5 in-class quizzes. The concepts from each project build on top of each other to culminate into the final drone obstacle race. Each project is performed in groups of three since that optimally balances enabling teamwork and giving a great experience to the students and enough time with the robots. The projects expose the students to the real-world challenges of integration, filtering and edge case handling of autonomy stack on aerial robots with real-time constraints. In-class quizzes help to strengthen mathematical concepts without the burden of large-scale exams. The course projects cover a wide range of topics from sensing, sensor fusion, planning, control, vision, integration and AI. The undergraduate and graduate versions of the course have slightly different project expectations which will be denoted by \textcolor{red}{UG} and \textcolor{red}{G} and are described next. The sensor suite used in \textcolor{red}{UG} and \textcolor{red}{G} versions involve RGBD (front + down-facing with IMU) and RGB (front + down-facing with IMU) respectively.

\textbf{Project 0:} Since the course is fast-paced and has high-expectations of time and effort commitment from students, there is a zeroth project that is due on the second day of class. This helps students gauge the difficulty level and their preparedness to take the course. The project consists of implementing a simple complementary filter for attitude estimation and setting up the Blender$^\text{\textregistered}$ quadrotor simulator to get familiar with the API. This project remains same for \textcolor{red}{UG} and \textcolor{red}{G}.

\textbf{Project 1:} The next project covers topics from sensor fusion to estimate attitude from a 6-DoF Inertial Measurement Unit (IMU) using Madgwick\cite{madgwick2010efficient} (\textcolor{red}{UG} and \textcolor{red}{G}) and Unscented Kalman Filters\cite{ukf} (\textcolor{red}{G}).

\textbf{Project 2:} In this project, students learn about path planning in a known map and also get exposed to the robot hardware to learn the API for path following using acceleration, velocity and position commands. The students implement A*/RRT* path planners in the \textcolor{red}{UG}/\textcolor{red}{G} versions respectively in the Blender$^\text{\textregistered}$ simulator. For the hardware experiments, the \textcolor{red}{UG} expects positional waypoint commands only whereas the the \textcolor{red}{G} expects both positional and velocity commands to follow polynomial trajectories.

\label{sec:project3}
\textbf{Project 3:} This project covers concepts of instance segmentation, sim2real transfer for neural networks and synthetic data generation. The goal is to detect known custom racing windows inspired from the AlphaPilot challenge\cite{foehn2022alphapilot}. The students generate synthetic data in Blender$^\text{\textregistered}$, use it to train a neural network (with no real data) to distinguish between windows. Then they use these detected windows to estimate their 3D pose to fly through them. 

\textbf{Project 4:} This project is inspired by real-world issues in search and rescue operations: to fly through unknown shaped gaps. This project is more open-ended than the previous ones, where the sensor suite can drastically change the approach to the solution. In previous offerings, students have solved this assignment using depth based clustering (in RGBD case) and optical flow based clustering (in RGB case) from concepts inspired from \cite{gapflyt, sanket2021morpheyes}.

\begin{figure*}[t!]
    \centering
    \includegraphics[width=0.85\textwidth]{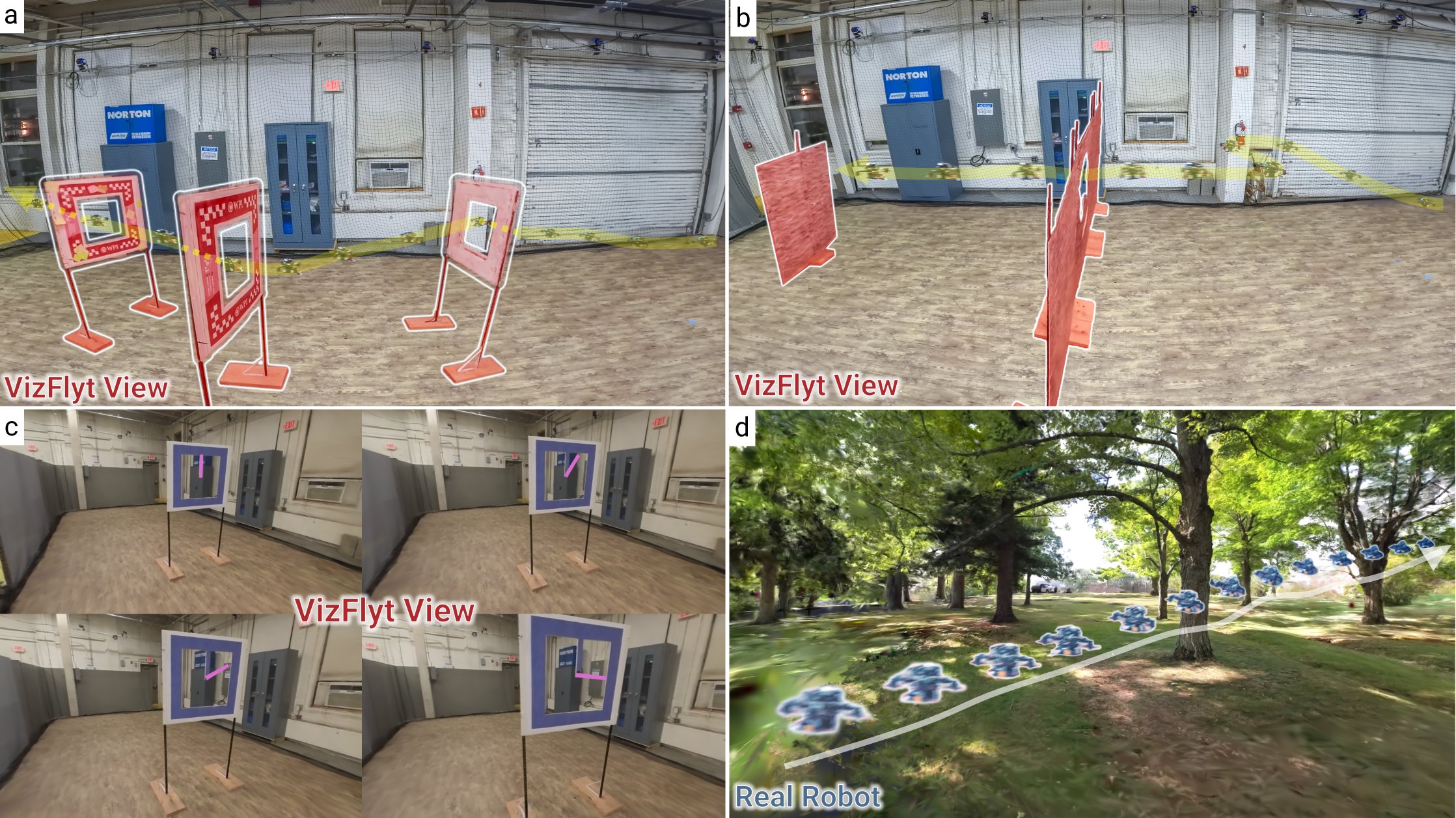}
    \caption{(a) Flight test for proposed project 3: flight through racing windows, (b)  Flight test for proposed project 4: flight through unknown gap, (c) Proposed project 5's dynamic window and (d) High-speed flight experiment. The red highlights denoted as \textit{VizFlyt} view show hallucinated cameras seen by the robot in (a) to (c). In (d), the entire scene is hallucinated with the real-robot overlaid.}
    \label{fig:nav_results}
\end{figure*}

\textbf{Project 5:} The final project culminates the learnings from all the previous projects into a final race through a tough obstacle course which also contains a dynamic window, which is a colored rectangular window with a clock like hand rotating at a fixed speed. The robots start from a starting position and navigate through the course in stages involving two racing windows, an unknown shaped gap and finally the dynamic window (Fig. \ref{fig:Overview}\textcolor{red}{c}). Each robot run is timed with the timer starting at takeoff and the timer stops the moment the robot passes through the dynamic window or collides with any part of the scene (in real-life such as nets or in virtual reality through our auto-grader). A leaderboard is maintained with the number of stages passed successfully and the time taken. The team with the lowest time and the most number of stages wins bragging rights and a trophy. 

\subsection{Grading and Evaluation}
Since the focus of the course is on real-world deployment of autonomy, each project is evaluated on live experiments. 
To enable efficient evaluation, we also develop an autograder system for live experiments based on collision detection from $\S$\ref{sec:collisioncheck} 3DGS (Fig. \ref{fig:collision_check}). The specific evaluation depends on factors like safety, speed of task completion and robustness to various factors. Furthermore, our system allows evaluation on various environmental conditions. This only takes a few seconds rather than hours to physically change the environmental settings for every team in the past iterations of our course.

\subsection{Logistics and Cost of Operation}
We also try to minimize the course logistics by streamlining the operation. \textit{VizFlyt} is centered around having a pose estimate for the robot to hallucinate vision based sensors. This can be achieved through an optical motion capture system (commonly found in aerial robotics research labs) or a much more cost-efficient Ultra-Wideband (UWB) position system like LoCo. This has a one time cost of 5-50K USD which remains operational for 10+ years with minimal maintenance. Each robot costs around 475USD (reducible to 300USD with basic controllers). The embedded computer would incur an additional cost. The radio transmitter used for safety kill switch costs about 95USD. During our \textit{VizFlyt} experiments and extensive stress testing over the last 8 months from nine lab members of all educational levels, we only broke three propellers, burnt two motors, three batteries and two 3D printed parts costing us a total of 125USD. Extrapolating this information, we estimate 62.5USD per group per year in maintenance/replacement costs. We also assume that the course takes two robots to run which will be replaced in their entirety every year. The cost to run the course for 30 students per semester in teams of three for 10 years for both semesters in the year including all the equipment and maintenance cost would be 110USD/student/offering including buying a Vicon$^\text{\textregistered}$ motion capture system (14 Vero V2.2 cameras) and 26.26USD/student/offering if a motion capture system exists. This is less than 1\% of the average student tuition. Lastly, due to changing regulations parts might need to be ordered from the same country, which are easily facilitated due to the hardware agnostic nature of our course. The students perform experiments first in Blender$^\text{\textregistered}$ simulation mated to 3DGS and then test their work out in the lab in their assigned lab slots. About 4hrs/team/week to run the hardware experiments is recommended from our experience. Furthermore, the open-source software stack and the hardware manuals would be maintained and updated yearly for long-term support.

\section{Experiments}
We test our \textit{VizFlyt} framework as a prototype for running the next generation of aerial robotics courses for vision-based autonomy through a series of experiments detailed next.


\subsection{Digital Twin Of The Test Scene}
We capture high-resolution (2K) images of real scenes from the previous offering of the course using an Insta360 GO 3 \cite{insta360}. We first use COLMAP\cite{colmap1, colmap2} to compute image poses and a sparse point cloud, which is used to train a high-fidelity model using Splatfacto \cite{NeRFStudio} with the default hyperparameters for 8000 iterations. It takes about $\approx5$ minutes to train a 3DGS model from 600 images given the image poses on the rendering PC.

\begin{figure*}[ht!]
    \centering
    \includegraphics[width=0.78\textwidth]{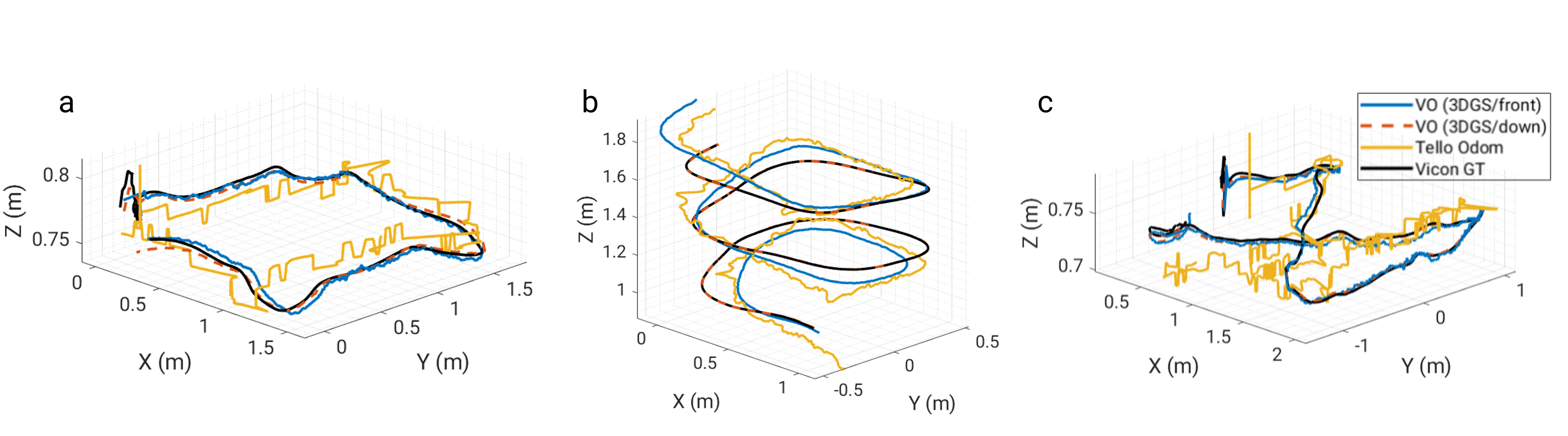}
    \caption{Comparison of various visual odometry methods on (a) square, (b) spiral and (c) lemniscate trajectories.}
    \label{fig:odom}
\end{figure*}

\subsection{Quadrotor Setup}
We use a custom-built quadrotor platform PeARWhippet160 for all our flight experiments. The robot has a Holybro Pix32 v6 flight controller running Arducopter v4.6.0-dev firmware and an ESP8266 module. The T-Motor F1507 3800KV motors are driven by the Hobbywing XRotor Micro 40A 6S BLHeli32 4-in-1 ESC. The main body frame is made from 2mm carbon fiber with 3D-printed mounts for additional components. The system is powered by a Tattu R-Line 4S 1300mAh LiPo battery. The all-up weight of the robot is about 400g (maximum thrust-to-weight ratio of 6.26:1) with a flight time of about 10 minutes. Six 9.5mm diameter passive Vicon$^\text{\textregistered}$ markers are placed in a unique 3D pattern on the robot to enable accurate 6D pose estimation.

\subsection{Flight Environment}
We test our \textit{VizFlyt} framework in the PeAR Washburn flying space which is a netted facility of dimension $11\,m \times 4.5\,m \times 3.65\,m$
 equipped with 14 Vicon$^\text{\textregistered}$ Vero V2.2 cameras running Tracker 4.0 software for accurate pose estimates.



\subsection{Prototype Course Experiment 1: Racing Window Navigation (Fig. \ref{fig:nav_results}\textcolor{red}{a})} 
This mimics project 3 of the proposed course from $\S$\ref{sec:project3}. The goal is to navigate through three racing windows of known size and shape as fast as possible. The windows are detected using a simple UNet-based\cite{unet} segmentation model trained on Blender$^\text{\textregistered}$ simulated images. The pose is recovered using \texttt{cv2.solvePnP} function from OpenCV, fed into a simple vector-field planner\cite{borenstein1991vector} mated to a PID controller. We achieved a max. speed of 1.65ms$^{-1}$ in our experiments.

\subsection{Prototype Course Experiment 2: Unknown Gap Navigation (Fig. \ref{fig:nav_results}\textcolor{red}{b})}
We designed project 4 of the course based on GapFlyt\cite{gapflyt}. We utilized the same strategy as GapFlyt and achieved similar statistics while navigating through an unknown gap.

\subsection{Prototype Course Experiment 3: Dynamic Window Navigation (Fig. \ref{fig:nav_results}\textcolor{red}{c}) }
The unique part of project 5 not covered in other parts is the dynamic window. Since we are using  3DGS, it does not support dynamic scene changes, i.e., moving the clock-like hand of the dynamic window. To enable this, we overlay a synthetic clock-like hand using temporally changing homography projections computed using the relative pose of the robot to the window \cite{malis_inria-00174036}. A similar approach generates the corresponding temporally changing occupancy grid for auto-grading. We use color thresholding combined with tracking and prediction of the hand using a linear model to fly through this window.

\subsection{Prototype Course Experiment 4: High-Speed Obstacle Avoidance (Fig. \ref{fig:nav_results}\textcolor{red}{d})}
Testing high-speed obstacle avoidance is challenging mainly due to the potential danger of damage to the robot and the environment. Our \textit{VizFlyt} framework can enable safe testing in such scenarios enabling future coursework and research on high-speed drone navigation. This also pushes the boundaries of our framework's utility by stress testing for latency and system update rate. We utilize the hallucinated depth images coupled to a potential field algorithm \cite{hwang1992potential} to navigate towards a goal direction while avoiding obstacles. The experiment is performed on an outdoor 3DGS forest environment and a max. speed of 5.3ms$^{-1}$ was achieved.



\subsection{Prototype Course Experiment 4: Visual Odometry}
To compare \textit{VizFlyt} with the previous course offering, we devise a simple Visual Odometry (VO)) experiment. A DJI Tello Edu is flown to execute three different trajectories: square, spiral and lemniscate. The ground truth poses were obtained from the Vicon$^\text{\textregistered}$ system. We hallucinate a front and a down-facing RGB camera. We compare the state-of-the-art DPVO \cite{teed2024deep} on our hallucinated camera images against the ground truth poses, the built-in odometry from DJI Tello Edu over the Absolute Trajectory Error (ATE) metric using the MATLAB implementation of \cite{Zhang18iros, sturm2012benchmark}. We obtain a maximum ATE error of 0.12m using hallucinated images on all the trajectories, this is 2.8$\times$ better than Tello's in-built odometry (Fig. \ref{fig:odom}). The results indicate a minimal sim-to-real gap and ensure that our framework can be used for teaching/researching VO methods similar to what datasets \cite{antonini2018blackbird, tartanair2020iros} offer but supplemented by HITL.   



\section{Discussion and Future Work}
With \textit{VizFlyt}, all the course projects were completed with lower effort, time and number of crashes leading to less frustrating and more efficient learning experience.  We also observed that using hallucinated photorealistic sensors provided a virtually identical learning experience to using real cameras on robots. Since our framework is adaptable, we can build experiments that are not easily possible with real sensors such as changing intrinsics and extrinsics of sensors, and placement of multiple sensors since they can be changed real-time and on-the-fly. Furthermore, testing over multiple scenarios is easier with our framework.

The conceptualization of this work started from the frustration built up as instructors of the course. However, during the development of the \textit{VizFlyt} framework, we noticed that such an approach is vital to push the boundaries of vision-based drone autonomy research further. This coupled with regulations of flying outside limits the capabilities of testing outdoors which can be now hallucinated. Furthermore, this forms a cycle of development that spurs from teaching to building a robust framework for testing, which in turn aids higher-quality research. This research enables better course content to train the future workforce. This self-reinforcing loop is favorable to society and forms an efficient teaching-research synergistic loop. 

For a future direction, it is simple to integrate adverse weather-based effects (through image translation networks \cite{vinod2021multi} or wavelet method for simulating fluids \cite{smoke_simulation}) into our hallucinated images to test autonomy software easily which is currently arduous. Note that, these only model the perception-centric sensor characteristics and not the physical effects on the robot. Furthermore, modeling and simulating commonly used depth-based sensors such as LiDAR, RADAR and Ultrasound is possible with minimal effort using the depth maps hallucinated by our work. This will enable pushing embodied AI further for aerial robots through data generation, testing and evaluation. 



\section{Conclusion}
The performance of autonomy software on aerial robots is bounded by their testing frameworks. This is even more true when students need to be trained on interdisciplinary concepts and algorithms for advanced autonomy with real-time constraints. Testing by deploying software on robots with obstacles in the scene not only makes it challenging but is time and resource-intensive due to frequent crashes which detracts students from learning. We presented \textit{VizFlyt}, a HITL framework to hallucinate photorealistic visual sensors in real-time to enable reliable, repeatable and flexible testing of drone autonomy software. We use \textit{VizFlyt} to propose a new curriculum to prepare the next generation of workforce in aerial robot autonomy. We test the proposed projects in our framework and showcase their efficacy. We believe \textit{VizFlyt} will not only enable efficient teaching but also aid aerial robotics research to push the boundaries of autonomy further.



\bibliographystyle{IEEEtran}

\bibliography{main}

\begin{thebibliography}{10}
\providecommand{\url}[1]{#1}
\csname url@samestyle\endcsname
\providecommand{\newblock}{\relax}
\providecommand{\bibinfo}[2]{#2}
\providecommand{\BIBentrySTDinterwordspacing}{\spaceskip=0pt\relax}
\providecommand{\BIBentryALTinterwordstretchfactor}{4}
\providecommand{\BIBentryALTinterwordspacing}{\spaceskip=\fontdimen2\font plus
\BIBentryALTinterwordstretchfactor\fontdimen3\font minus \fontdimen4\font\relax}
\providecommand{\BIBforeignlanguage}[2]{{%
\expandafter\ifx\csname l@#1\endcsname\relax
\typeout{** WARNING: IEEEtran.bst: No hyphenation pattern has been}%
\typeout{** loaded for the language `#1'. Using the pattern for}%
\typeout{** the default language instead.}%
\else
\language=\csname l@#1\endcsname
\fi
#2}}
\providecommand{\BIBdecl}{\relax}
\BIBdecl

\bibitem{search_rescue}
K.~McGuire, C.~De~Wagter, K.~Tuyls, H.~Kappen, and G.~C. de~Croon, ``Minimal navigation solution for a swarm of tiny flying robots to explore an unknown environment,'' \emph{Science Robotics}, vol.~4, no.~35, p. eaaw9710, 2019.

\bibitem{farming}
P.~Tokekar, J.~V. Hook, D.~Mulla, and V.~Isler, ``Sensor planning for a symbiotic uav and ugv system for precision agriculture,'' \emph{IEEE Transactions on Robotics}, vol.~32, no.~6, pp. 1498--1511, 2016.

\bibitem{inspection}
T.~Nguyen, T.~Ozaslan, I.~D. Miller, J.~Keller, G.~Loianno, C.~J. Taylor, D.~D. Lee, V.~Kumar, J.~H. Harwood, and J.~Wozencraft, ``U-net for mav-based penstock inspection: an investigation of focal loss in multi-class segmentation for corrosion identification,'' \emph{arXiv preprint arXiv:1809.06576}, 2018.

\bibitem{pirdone}
I.~Brand, J.~Roy, A.~Ray, J.~Oberlin, and S.~Oberlix, ``Pidrone: An autonomous educational drone using raspberry pi and python,'' in \emph{2018 IEEE/RSJ International Conference on Intelligent Robots and Systems (IROS)}, 2018, pp. 1--7.

\bibitem{mit_course}
L.~Carlone, K.~Khosoussi, V.~Tzoumas, G.~Habibi, M.~Ryll, R.~Talak, J.~Shi, and P.~Antonante, ``Visual navigation for autonomous vehicles: An open-source hands-on robotics course at mit,'' in \emph{2022 IEEE Integrated STEM Education Conference (ISEC)}.\hskip 1em plus 0.5em minus 0.4em\relax IEEE, 2022, pp. 177--184.

\bibitem{penn_course}
V.~Kumar, ``Robotics: Aerial robotics,'' now Archived.

\bibitem{tum_course}
\BIBentryALTinterwordspacing
J.~Sturm and D.~Cremers, ``Autonomous navigation for flying robots.'' [Online]. Available: \url{https://cvg.cit.tum.de/teaching/ss2015/autonavx}
\BIBentrySTDinterwordspacing

\bibitem{umd_course}
\BIBentryALTinterwordspacing
N.~J. Sanket and C.~D. Singh, ``Enae 788m: Hands on autonomous aerial robotics.'' [Online]. Available: \url{https://prg.cs.umd.edu/enae788m}
\BIBentrySTDinterwordspacing

\bibitem{rbe595}
\BIBentryALTinterwordspacing
{Nitin J. Sanket}, ``Hands-on autonomous aerial robotics.'' [Online]. Available: \url{https://pear.wpi.edu/teaching/rbe595/fall2023.html}
\BIBentrySTDinterwordspacing

\bibitem{meam_penn_course}
\BIBentryALTinterwordspacing
G.~Loianno, J.~Paulos, and K.~Daniilidis, ``Meam 620: Robotics.'' [Online]. Available: \url{https://alliance.seas.upenn.edu/~meam620/wiki/}
\BIBentrySTDinterwordspacing

\bibitem{osu}
\BIBentryALTinterwordspacing
{Ohio State University}, ``Avn 2400,'' n.d., [Online; accessed 15-Sep-2024]. [Online]. Available: \url{https://www.catalogs.ohio.edu/preview_course_nopop.php?catoid=82&coid=440277}
\BIBentrySTDinterwordspacing

\bibitem{osu1}
\BIBentryALTinterwordspacing
``Avn 2400: Ohio state universtiy,'' n.d., [Online; accessed 15-Sep-2024]. [Online]. Available: \url{https://syllabi.engineering.osu.edu/syllabi/aviatn_2401}
\BIBentrySTDinterwordspacing

\bibitem{mambo}
\BIBentryALTinterwordspacing
``Parrot mambo.'' [Online]. Available: \url{https://www.parrot.com/en/support/documentation/mambo-range}
\BIBentrySTDinterwordspacing

\bibitem{djitello}
\BIBentryALTinterwordspacing
``Dji tello.'' [Online]. Available: \url{https://store.dji.com/product/tello-edu}
\BIBentrySTDinterwordspacing

\bibitem{rollingspider}
\BIBentryALTinterwordspacing
``Parrot rolling spider.'' [Online]. Available: \url{https://www.parrot.com/en/support/documentation/rolling-spider}
\BIBentrySTDinterwordspacing

\bibitem{flightmare}
Y.~Song, S.~Naji, E.~Kaufmann, A.~Loquercio, and D.~Scaramuzza, ``Flightmare: A flexible quadrotor simulator,'' in \emph{Conference on Robot Learning}.\hskip 1em plus 0.5em minus 0.4em\relax PMLR, 2021, pp. 1147--1157.

\bibitem{Gazebo}
C.~Aguero, N.~Koenig, I.~Chen, H.~Boyer, S.~Peters, J.~Hsu, B.~Gerkey, S.~Paepcke, J.~Rivero, J.~Manzo, E.~Krotkov, and G.~Pratt, ``Inside the virtual robotics challenge: Simulating real-time robotic disaster response,'' \emph{Automation Science and Engineering, IEEE Transactions on}, vol.~12, no.~2, pp. 494--506, April 2015.

\bibitem{IsaacSim}
\BIBentryALTinterwordspacing
``{NVIDIA Isaac Sim}.'' [Online]. Available: \url{https://developer.nvidia.com/isaac-sim}
\BIBentrySTDinterwordspacing

\bibitem{dimmig2024survey}
C.~A. Dimmig, G.~Silano, K.~McGuire, C.~Gabellieri, W.~H{\v{s}}nig, J.~Moore, and M.~Kobilarov, ``Survey of simulators for aerial robots: An overview and in-depth systematic comparisons,'' \emph{IEEE Robotics \& Automation Magazine}, 2024.

\bibitem{agilicious}
\BIBentryALTinterwordspacing
P.~Foehn, E.~Kaufmann, A.~Romero, R.~Penicka, S.~Sun, L.~Bauersfeld, T.~Laengle, G.~Cioffi, Y.~Song, A.~Loquercio, and D.~Scaramuzza, ``Agilicious: Open-source and open-hardware agile quadrotor for vision-based flight,'' \emph{Science Robotics}, vol.~7, no.~67, Jun. 2022. [Online]. Available: \url{http://dx.doi.org/10.1126/scirobotics.abl6259}
\BIBentrySTDinterwordspacing

\bibitem{FlightGoggles_MIT}
\BIBentryALTinterwordspacing
W.~Guerra, E.~Tal, V.~Murali, G.~Ryou, and S.~Karaman, ``Flightgoggles: Photorealistic sensor simulation for perception-driven robotics using photogrammetry and virtual reality,'' in \emph{2019 IEEE/RSJ International Conference on Intelligent Robots and Systems (IROS)}.\hskip 1em plus 0.5em minus 0.4em\relax IEEE, Nov. 2019. [Online]. Available: \url{http://dx.doi.org/10.1109/IROS40897.2019.8968116}
\BIBentrySTDinterwordspacing

\bibitem{song2023learning}
Y.~Song, K.~Shi, R.~Penicka, and D.~Scaramuzza, ``Learning perception-aware agile flight in cluttered environments,'' in \emph{2023 IEEE International Conference on Robotics and Automation (ICRA)}.\hskip 1em plus 0.5em minus 0.4em\relax IEEE, 2023, pp. 1989--1995.

\bibitem{kerbl3Dgaussians}
\BIBentryALTinterwordspacing
B.~Kerbl, G.~Kopanas, T.~Leimk{\"u}hler, and G.~Drettakis, ``3d gaussian splatting for real-time radiance field rendering,'' \emph{ACM Transactions on Graphics}, vol.~42, no.~4, July 2023. [Online]. Available: \url{https://repo-sam.inria.fr/fungraph/3d-gaussian-splatting/}
\BIBentrySTDinterwordspacing

\bibitem{yiannis}
\BIBentryALTinterwordspacing
Y.~Aloimonos and N.~J. Sanket, ``Cmsc828t: Vision, planning and control in aerial robotics.'' [Online]. Available: \url{https://cmsc828t.github.io/}
\BIBentrySTDinterwordspacing

\bibitem{fire}
\BIBentryALTinterwordspacing
N.~J. Sanket, ``Fire198: Fire semester 2 autonomous unmanned systems.'' [Online]. Available: \url{https://umdausfire.github.io/teaching/fire198/spring2022.html}
\BIBentrySTDinterwordspacing

\bibitem{foehn2022alphapilot}
P.~Foehn, D.~Brescianini, E.~Kaufmann, T.~Cieslewski, M.~Gehrig, M.~Muglikar, and D.~Scaramuzza, ``Alphapilot: Autonomous drone racing,'' \emph{Autonomous Robots}, vol.~46, no.~1, pp. 307--320, 2022.

\bibitem{motorsport_concept}
\BIBentryALTinterwordspacing
{ASPIRE's Executive Director, Dr. Tom McCarthy}, ``The motorsport concept: Building an autonomous mobility ecosystem,'' n.d., [Online; accessed 15-Sep-2024]. [Online]. Available: \url{https://a2rl.io}
\BIBentrySTDinterwordspacing

\bibitem{drl_website}
\BIBentryALTinterwordspacing
``Drone racing league (drl),'' n.d., [Online; accessed 15-Sep-2024]. [Online]. Available: \url{https://www.drl.io/}
\BIBentrySTDinterwordspacing

\bibitem{vicon}
\BIBentryALTinterwordspacing
``Vicon motion capture system.'' [Online]. Available: \url{https://www.vicon.com/}
\BIBentrySTDinterwordspacing

\bibitem{bitcraze_loco}
\BIBentryALTinterwordspacing
``Loco positioning system.'' [Online]. Available: \url{https://www.bitcraze.io/documentation/system/positioning/loco-positioning-system/}
\BIBentrySTDinterwordspacing

\bibitem{DBLP:conf/icra/PfrommerSDC17}
\BIBentryALTinterwordspacing
B.~Pfrommer, N.~Sanket, K.~Daniilidis, and J.~Cleveland, ``Penncosyvio: {A} challenging visual inertial odometry benchmark,'' in \emph{2017 {IEEE} International Conference on Robotics and Automation, {ICRA} 2017, Singapore, Singapore, May 29 - June 3, 2017}, 2017, pp. 3847--3854. [Online]. Available: \url{https://doi.org/10.1109/ICRA.2017.7989443}
\BIBentrySTDinterwordspacing

\bibitem{lighthouse}
\BIBentryALTinterwordspacing
``Lighthouse positioning system.'' [Online]. Available: \url{https://www.bitcraze.io/documentation/system/positioning/ligthouse-positioning-system/}
\BIBentrySTDinterwordspacing

\bibitem{ros2}
\BIBentryALTinterwordspacing
S.~Macenski, T.~Foote, B.~Gerkey, C.~Lalancette, and W.~Woodall, ``Robot operating system 2: Design, architecture, and uses in the wild,'' \emph{Science Robotics}, vol.~7, no.~66, p. eabm6074, 2022. [Online]. Available: \url{https://www.science.org/doi/abs/10.1126/scirobotics.abm6074}
\BIBentrySTDinterwordspacing

\bibitem{madgwick2010efficient}
S.~Madgwick \emph{et~al.}, ``An efficient orientation filter for inertial and inertial/magnetic sensor arrays,'' \emph{Report x-io and University of Bristol (UK)}, vol.~25, pp. 113--118, 2010.

\bibitem{ukf}
E.~A. Wan and R.~Van Der~Merwe, ``The unscented kalman filter for nonlinear estimation,'' in \emph{Proceedings of the IEEE 2000 adaptive systems for signal processing, communications, and control symposium (Cat. No. 00EX373)}.\hskip 1em plus 0.5em minus 0.4em\relax Ieee, 2000, pp. 153--158.

\bibitem{gapflyt}
N.~J. Sanket, C.~D. Singh, K.~Ganguly, C.~Ferm\"uller, and Y.~Aloimonos, ``Gapflyt: Active vision based minimalist structure-less gap detection for quadrotor flight,'' \emph{IEEE Robotics and Automation Letters}, vol.~3, no.~4, pp. 2799--2806, Oct 2018.

\bibitem{sanket2021morpheyes}
N.~J. Sanket, C.~D. Singh, V.~Asthana, C.~Ferm{\"u}ller, and Y.~Aloimonos, ``Morpheyes: Variable baseline stereo for quadrotor navigation,'' in \emph{2021 IEEE International Conference on Robotics and Automation (ICRA)}.\hskip 1em plus 0.5em minus 0.4em\relax IEEE, 2021, pp. 413--419.

\bibitem{insta360}
\BIBentryALTinterwordspacing
``Insta360 go 3.'' [Online]. Available: \url{https://store.insta360.com/product/go-3}
\BIBentrySTDinterwordspacing

\bibitem{colmap1}
J.~L. Sch\"{o}nberger and J.-M. Frahm, ``Structure-from-motion revisited,'' in \emph{Conference on Computer Vision and Pattern Recognition (CVPR)}, 2016.

\bibitem{colmap2}
J.~L. Sch\"{o}nberger, E.~Zheng, M.~Pollefeys, and J.-M. Frahm, ``Pixelwise view selection for unstructured multi-view stereo,'' in \emph{European Conference on Computer Vision (ECCV)}, 2016.

\bibitem{NeRFStudio}
\BIBentryALTinterwordspacing
M.~Tancik, E.~Weber, E.~Ng, R.~Li, B.~Yi, T.~Wang, A.~Kristoffersen, J.~Austin, K.~Salahi, A.~Ahuja, D.~Mcallister, J.~Kerr, and A.~Kanazawa, ``Nerfstudio: A modular framework for neural radiance field development,'' in \emph{Special Interest Group on Computer Graphics and Interactive Techniques Conference Conference Proceedings}, ser. SIGGRAPH ’23.\hskip 1em plus 0.5em minus 0.4em\relax ACM, Jul. 2023. [Online]. Available: \url{http://dx.doi.org/10.1145/3588432.3591516}
\BIBentrySTDinterwordspacing

\bibitem{unet}
O.~Ronneberger, P.~Fischer, and T.~Brox, ``U-net: Convolutional networks for biomedical image segmentation,'' in \emph{Medical image computing and computer-assisted intervention--MICCAI 2015: 18th international conference, Munich, Germany, October 5-9, 2015, proceedings, part III 18}.\hskip 1em plus 0.5em minus 0.4em\relax Springer, 2015, pp. 234--241.

\bibitem{borenstein1991vector}
J.~Borenstein, Y.~Koren \emph{et~al.}, ``The vector field histogram-fast obstacle avoidance for mobile robots,'' \emph{IEEE transactions on robotics and automation}, vol.~7, no.~3, pp. 278--288, 1991.

\bibitem{malis_inria-00174036}
\BIBentryALTinterwordspacing
E.~Malis and M.~Vargas, ``{Deeper understanding of the homography decomposition for vision-based control},'' {INRIA}, Research Report RR-6303, 2007. [Online]. Available: \url{https://inria.hal.science/inria-00174036}
\BIBentrySTDinterwordspacing

\bibitem{hwang1992potential}
Y.~K. Hwang, N.~Ahuja \emph{et~al.}, ``A potential field approach to path planning.'' \emph{IEEE transactions on robotics and automation}, vol.~8, no.~1, pp. 23--32, 1992.

\bibitem{teed2024deep}
Z.~Teed, L.~Lipson, and J.~Deng, ``Deep patch visual odometry,'' \emph{Advances in Neural Information Processing Systems}, vol.~36, 2024.

\bibitem{Zhang18iros}
Z.~Zhang and D.~Scaramuzza, ``A tutorial on quantitative trajectory evaluation for visual(-inertial) odometry,'' in \emph{IEEE/RSJ Int. Conf. Intell. Robot. Syst. (IROS)}, 2018.

\bibitem{sturm2012benchmark}
J.~Sturm, N.~Engelhard, F.~Endres, W.~Burgard, and D.~Cremers, ``A benchmark for the evaluation of rgb-d slam systems,'' in \emph{2012 IEEE/RSJ international conference on intelligent robots and systems}.\hskip 1em plus 0.5em minus 0.4em\relax IEEE, 2012, pp. 573--580.

\bibitem{antonini2018blackbird}
\BIBentryALTinterwordspacing
A.~Antonini, W.~Guerra, V.~Murali, T.~Sayre-McCord, and S.~Karaman, ``The blackbird dataset: A large-scale dataset for uav perception in aggressive flight,'' in \emph{2018 International Symposium on Experimental Robotics (ISER)}, 2018. [Online]. Available: \url{https://doi.org/10.1007/978-3-030-33950-0_12}
\BIBentrySTDinterwordspacing

\bibitem{tartanair2020iros}
W.~Wang, D.~Zhu, X.~Wang, Y.~Hu, Y.~Qiu, C.~Wang, Y.~Hu, A.~Kapoor, and S.~Scherer, ``Tartanair: A dataset to push the limits of visual slam,'' 2020.

\bibitem{vinod2021multi}
V.~Vinod, K.~R. Prabhakar, R.~V. Babu, and A.~Chakraborty, ``Multi-domain conditional image translation: Translating driving datasets from clear-weather to adverse conditions,'' in \emph{Proceedings of the IEEE/CVF International Conference on Computer Vision}, 2021, pp. 1571--1582.

\bibitem{smoke_simulation}
T.~Kim, N.~Th{\"u}rey, D.~James, and M.~Gross, ``Wavelet turbulence for fluid simulation,'' \emph{ACM Transactions on Graphics (TOG)}, vol.~27, no.~3, pp. 1--6, 2008.

\end{thebibliography}

\end{document}